\documentclass{article}

 \usepackage[main, final]{neurips_2026}

\usepackage[utf8]{inputenc} 
\usepackage[T1]{fontenc}    
\usepackage{hyperref}       
\usepackage{url}            
\usepackage{booktabs}       
\usepackage{amsfonts}       
\usepackage{nicefrac}       
\usepackage{microtype}      
\usepackage{xcolor}         
\usepackage[pdftex]{graphicx}
\usepackage{cleveref}
\usepackage{wrapfig}

\title{CityRAG: Stepping Into a City via Spatially-Grounded Video Generation}

\author{
\textbf{
Gene Chou$^{1,2}$ \quad
Charles Herrmann$^{1}$ \quad
Kyle Genova$^{1}$ \quad
Boyang Deng$^{3}$ \quad
Songyou Peng$^{1}$} \\
\textbf{
Bharath Hariharan$^{2}$ \quad
Jason Y. Zhang$^{1}$ \quad
Noah Snavely$^{1,2}$ \quad
Philipp Henzler$^{1}$} \\
\\[-0.5ex]
$^{1}$Google \quad
$^{2}$Cornell University \quad
$^{3}$Stanford University
}

\begin{document}

\maketitle

\begin{abstract}
We address the problem of generating a 3D-consistent, navigable environment that is spatially grounded: a simulation of a real location. Existing video generative models can produce a plausible sequence that is consistent with a text (T2V) or image (I2V) prompt. However, the capability to reconstruct the real world under arbitrary weather conditions and dynamic object configurations is essential for downstream applications including autonomous driving and robotics simulation. To this end, we present CityRAG, a video generative model that leverages large corpora of geo-registered data as context to ground generation to the physical scene, while maintaining learned priors for complex motion and appearance changes. CityRAG relies on temporally unaligned training data, which teaches the model to semantically disentangle the underlying scene from its transient attributes. Our experiments demonstrate that CityRAG can generate coherent minutes-long, physically grounded video sequences, maintain weather and lighting conditions over thousands of frames, achieve loop closure, and navigate complex trajectories to reconstruct real-world geography. See our \href{https://cityrag.github.io/}{\textcolor{blue}{\underline{website}}} for video playback.
\end{abstract}    
\section{Introduction}
\label{sec:intro}

\begin{figure}[h]
  \centering
  \includegraphics[width=\textwidth]{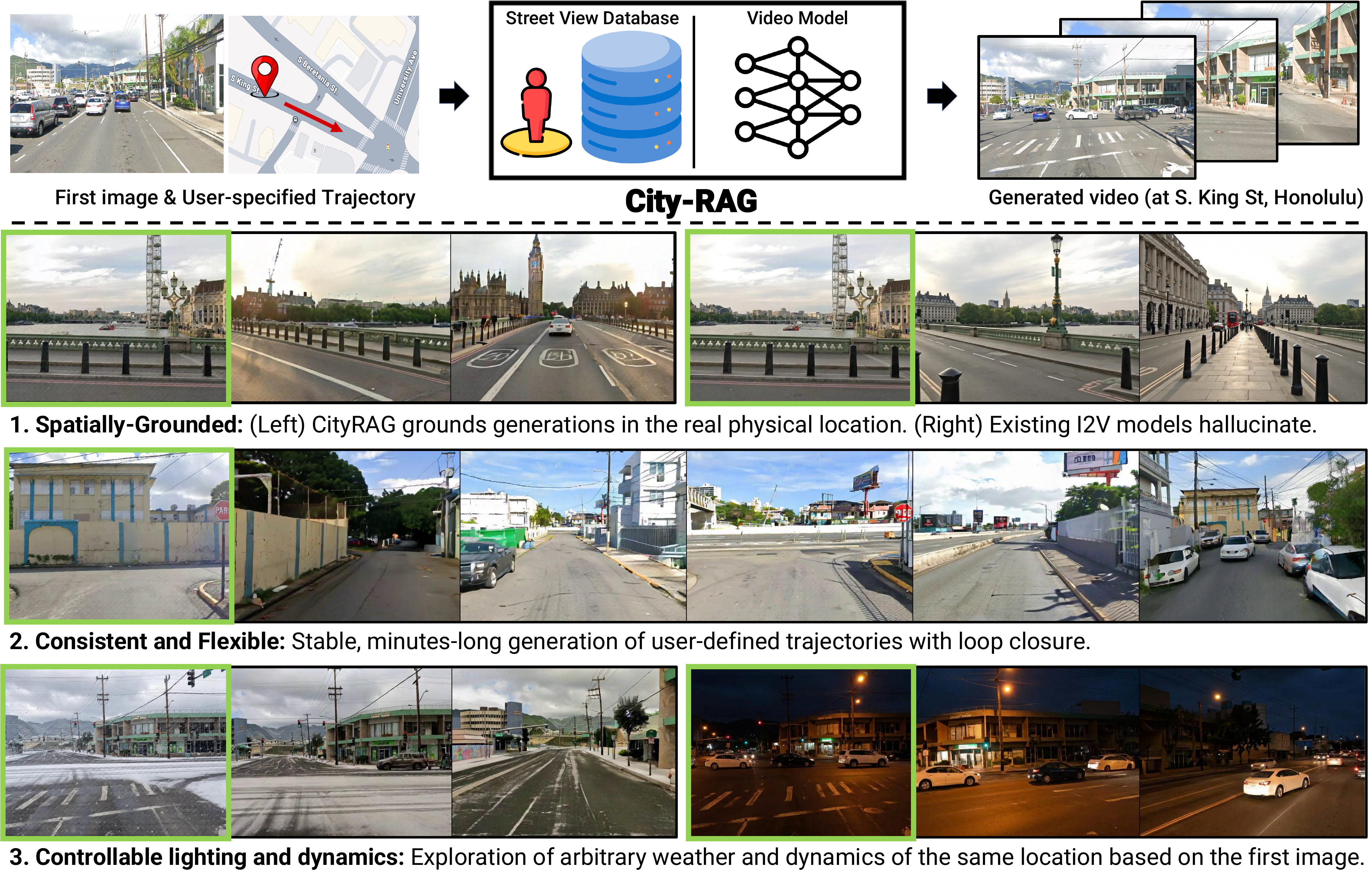}
  \caption{
  CityRAG generates minutes-long, spatially grounded video sequences that 1) Render real buildings, traffic lights, and roads of a city. From an image taken on the Westminster Bridge in London, CityRAG generates Big Ben and the Houses of Parliament as the viewpoint rotates left, while Veo \citep{google2025veo3} hallucinates. 2) Follow a user-defined path and perform loop closure after generating a thousand frames. 3) Are initialized from a first image and respects its weather conditions and dynamic objects. \textit{Top:} The Westminster Bridge, London. \textit{Middle:} Calle Quiñones St, San Juan. \textit{Bottom:} S King St, Honolulu. Starting views are labeled with green bounding boxes.
  }
  \label{fig:teaser}
  \vspace{-2em}
\end{figure}

Imagine pulling up a photo of New York City, taken from the intersection of 42nd Street and 5th Avenue. Then, stepping into the image and walking toward the Empire State Building.
Although the landmark is not visible in the input photo, as the virtual camera moves south the whole of the city---the roads, traffic lights, shops, fire hydrants, the Empire State Building itself---perfectly match the geographic layout of the real world. 
Furthermore, the environment preserves the specific weather conditions of the photo (a light drizzle around 2pm) and its elements come to life: a taxi completes its turn and a man in a blazer continues walking.
In other words, the city is not a pure AI hallucination, but instead matches the real world; namely, the very world pictured in the input photo.

Such a capability would unlock applications in virtual tourism, gaming, and simulation for autonomous driving and robotics. 
For example, researchers could transform a snapshot of a snowstorm into a high-fidelity simulation to train self-driving cars, rather than driving thousands of miles in dangerous conditions \citep{waymo2026worldmodel}. 
Specialized robots could be trained to adapt to a specific environment, such as a factory, and learn to avoid transient objects like people and boxes while navigating around the corners \citep{gao2026dreamdojo}.

In this paper, we address the problem of generating a 3D-consistent, navigable environment that respects both the transient attributes of a first image condition, such as weather and pedestrians, and the static attributes derived from geospatial conditions, which take the form of pre-collected, geo-registered video frames, such as buildings and roads. Specifically, we focus on the domain of Street View for its dense coverage and semantic cues of the arrangement of static and dynamic elements. This allows us to ground generation in real-world environments. 

Achieving this requires querying and incorporating external context on-the-fly, a task that is difficult for existing approaches.
The dominant paradigm for generative models prioritizes scalability \citep{peebles2023scalable,blattmann2023stable} and thus relies on abundant and easily accessible data for conditioning, such as a text prompt or an image. But this approach cannot integrate external knowledge about the world during inference. 
On the other hand, non-generative 3D representations like NeRFs \citep{mildenhall2021nerf} require dense captures of the exact moment and lack the capacity to produce realistic motion or complex appearance changes.

To this end, we propose CityRAG, a video generative model that leverages large corpora of geo-registered data as context to guarantee fidelity to the scene, while maintaining learned priors for complex motion and appearance changes.

\begin{figure}[t!]
  \centering
  \includegraphics[width=\textwidth]{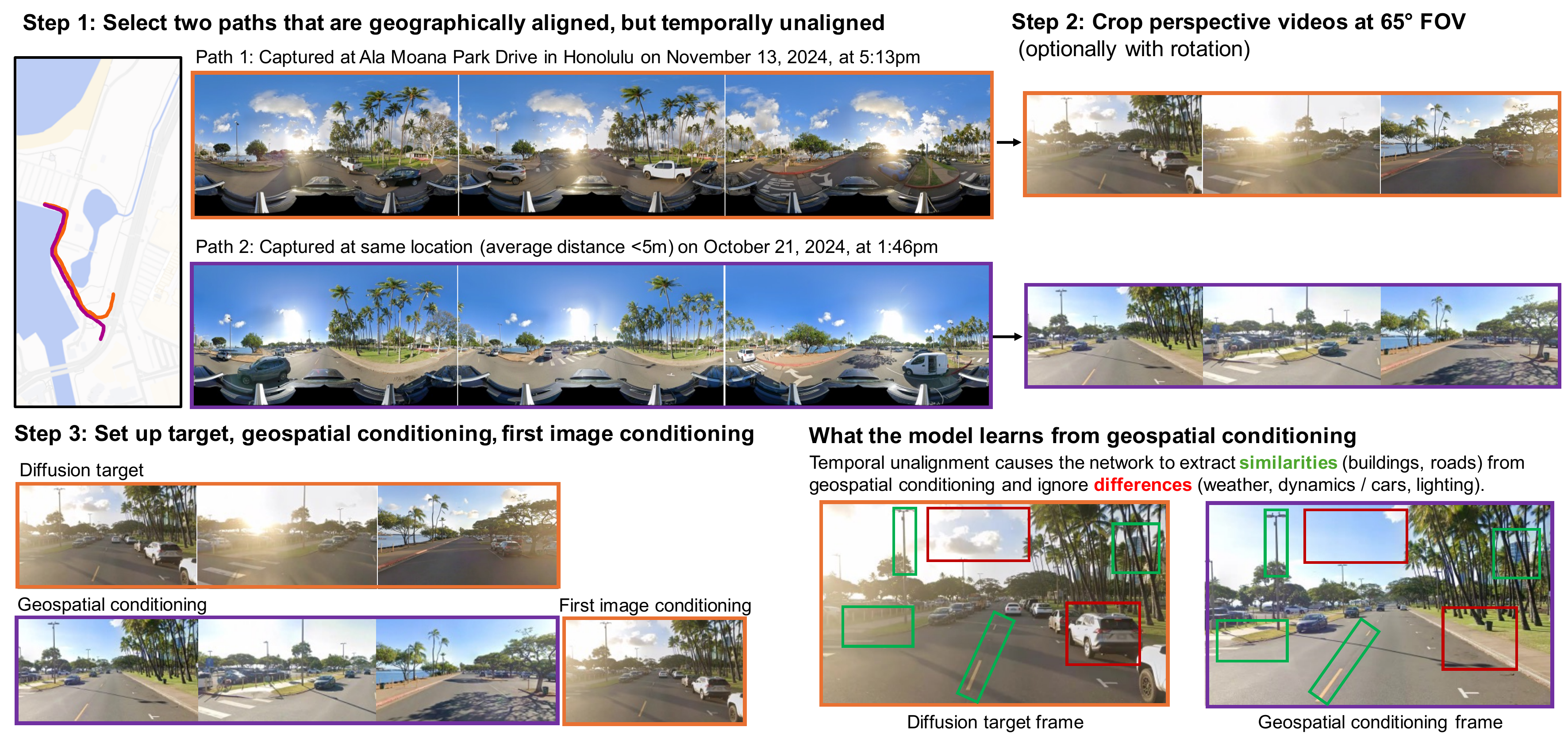}
  \caption{ \textbf{Training data pipeline.}
  We use Street View data in the form of panoramas. We create a training pair if there is a continuous path where there exists 2 sets of captures at different times (e.g., morning vs. afternoon) but with an average distance < 5 meters, so the model learns to disentangle static and transient attributes, e.g., roads and buildings (green box) vs. weather and cars (red box). \\
  }
  \label{fig:data}
  \vspace{-2em}
\end{figure}

Starting from an input image, CityRAG retrieves a multi-view ``memory'' of the location and injects it through a dedicated branch of attention layers. 
This architecture teaches the model to extract two distinct sets of information: transients from the image, such as lighting and dynamic objects, and statics from the ``memory,'' such as buildings and roads. Through a carefully designed data-driven strategy, CityRAG learns to decouple and recombine these attributes, visualized in \cref{fig:data} and \cref{fig:arch}.

First, we curate a dataset of paired Street View videos that capture the same physical location at different times (e.g., morning vs.\ sunset) (\cref{sec:data}). 
This provides the data required for a model to semantically distinguish between static and transient attributes. 
Specifically, we collect a total of 5.5M Street View panoramas and their poses across 10 cities. These paired sequences allow a model to observe the same streets under diverse illumination and traffic conditions.

Second, we finetune a state-of-the-art I2V model, Wan 2.1 \citep{wan2025}, on the paired data (\cref{sec:arch}).
While the pretrained model adheres to a first image condition, it lacks context beyond the immediate field of view. To address this, we introduce a training strategy that uses temporally unaligned frames (images of the same location captured at different times) as a structural anchor. By forcing the model to derive a static layout from morning frames to reconstruct a scene at night, we decouple permanent geometry from transient environmental conditions.

During inference, given an input image and defined trajectory, CityRAG retrieves videos from the vicinity to serve as a reliable prior for the scene's identity (\cref{sec:inference}). 
As the model learns to faithfully reconstruct the buildings and roads, generated videos remain consistent across independent and sequential samples, even without being trained for autoregression. 
The result is a model capable of generating minutes-long, 3D-consistent walkthroughs that simulate realistic motion of cars and pedestrians in a user's image while preserving the geography of the real physical location. 

We evaluate our approach via a variety of metrics, testing scenes, and baselines. 
We show that our approach demonstrates strong 3D understanding of the underlying scene, disentangles dynamic and static elements without heuristics, and generates realistic sequences across diverse settings. 

\section{Related Works}
\label{sec:related works}

\subsection{Video Generative Models}

Popular formulations for video generation include text-to-video (T2V) \citep{singer2022makeavideo,yang2024cogvideox} and image-to-video (I2V) \citep{sora, blattmann2023stable, BarTal2024LumiereAS} generation due to their scalability, and they can then be finetuned based on the requirements of downstream applications. Our application requires long-term consistency, pose control, and integration of external context.

\noindent\textbf{Long-term consistency.}
Works in long-context or autoregressive generation \citep{chen2025diffusion,krea_realtime_14b,song2025historyguidedvideodiffusion,zhang2025framepack,cai2025moc,xiao2025worldmemlongtermconsistentworld,huang2025selfforcing} maintain consistency by balancing computational efficiency and storing past samples. 
Another line of work creates an explicit memory like point clouds \citep{wu2025spmem,gu2025das,ren2025gen3c,Yu_2025_ICCV}. 
However, these works rarely show the capacity to generate minutes-long videos without significant degradation, and have an orthogonal focus to our work. CityRAG retrieves external context for grounding, rather than past samples, to maintain consistency.

\noindent\textbf{Pose-conditioning.}
Pose-conditioned models \citep{ren2025gen3c, bahmani2025ac3d, guo2023animatediff,vanhoorick2024gcd, wang2024motionctrl, tung2024megascenes, zhou2025stable} finetune a base generative model on camera poses, often in the form of camera parameters or warping and inpainting. They rely on generative priors of video models to remain temporally consistent and hallucinate plausible sequences while providing control. We similarly condition our model on camera extrinsics, but additionally adhere to large-scale real-world grounding.

\noindent\textbf{Using additional context.}
Reference-to-video (R2V) \citep{chen2025videoalchemist, wei2024dreamvideo, wang2024customvideo} and video-to-video (V2V) \citep{esser2023structure, geyer2023tokenflow, wu2024fairy, liang2024flowvid, ku2024anyv2v, zhou2025stable, liang2024looking, fu2025plenoptic} models are conceptually closer to our goal. 
But neither has shown 3D-awareness and both require strict adherence to the reference video. 
A few works experiment with conditioning without strict adherence. For example, LooseControl \citep{bhatloosecontrol2023} enables boundary control and scene editing with sparse depth maps. KFC-W \citep{chou2024kfcw} generates a 3D-consistent trajectory of a scene from random internet photos. However, none of them address a similar problem setting as ours. 


\subsection{Retrieval-Augmented Generation (RAG)}
RAG has been shown to mitigate hallucination and ground model outputs in external knowledge \citep{lewis2020rag}. Recently, this framework has been applied to visual generative models to enhance fidelity and realism. RealRAG \citep{lyu2025realrag} improves text-to-image synthesis by retrieving real-world reference images to fill in knowledge gaps during generation. MotionRAG \citep{zhu2025motionrag} retrieves video clips to provide demonstrations of motion. 
In a similar vein, our work retrieves geo-registered data to ground video generation in the real world.

\section{Method}
\label{sec:method}

\subsection{Data}
\label{sec:data}
With explicit permission from Google, we collect Street View data from Google Maps across 10 diverse cities scattered across the globe: Paris, Athens, Anchorage, Hyderabad, Philadelphia, San Francisco, San Juan, Honolulu, London, and Sao Paolo. 
Importantly, all sensitive information, such as license plates and faces, are blurred prior to collection.

The data is in the form of panoramas and their associated poses in the Earth-Centered, Earth-Fixed (ECEF) coordinate system. Thus, the poses are in metric scale (meters) and consistent across all cities. 
We sample with density roughly equivalent to 10 FPS. 
In addition, we sample captures of the same streets at different times, when available. We collected a total of 5.5M pano-pose pairs. 

Then, we group all panoramas by their trajectories and time of capture. 
We create a training pair if we find a continuous path in a city of length N where there exists 2 sets of panoramas located along the same path with an average distance threshold smaller than $\epsilon$ meters, but captured at different times (e.g, different dates, or even morning vs. afternoon of the same day). We set N=73 the number of frames sampled for training, and $\epsilon=5$.
After filtering for these pairs, we obtain a total of 1.3M panoramas for training and a few thousand held out for testing. We show an example pair in \cref{fig:data}. 

\subsection{Architecture}
\label{sec:arch}
\begin{wrapfigure}{r}{0.6\textwidth} 
\vspace{-2em}
\includegraphics[width=0.6\textwidth]{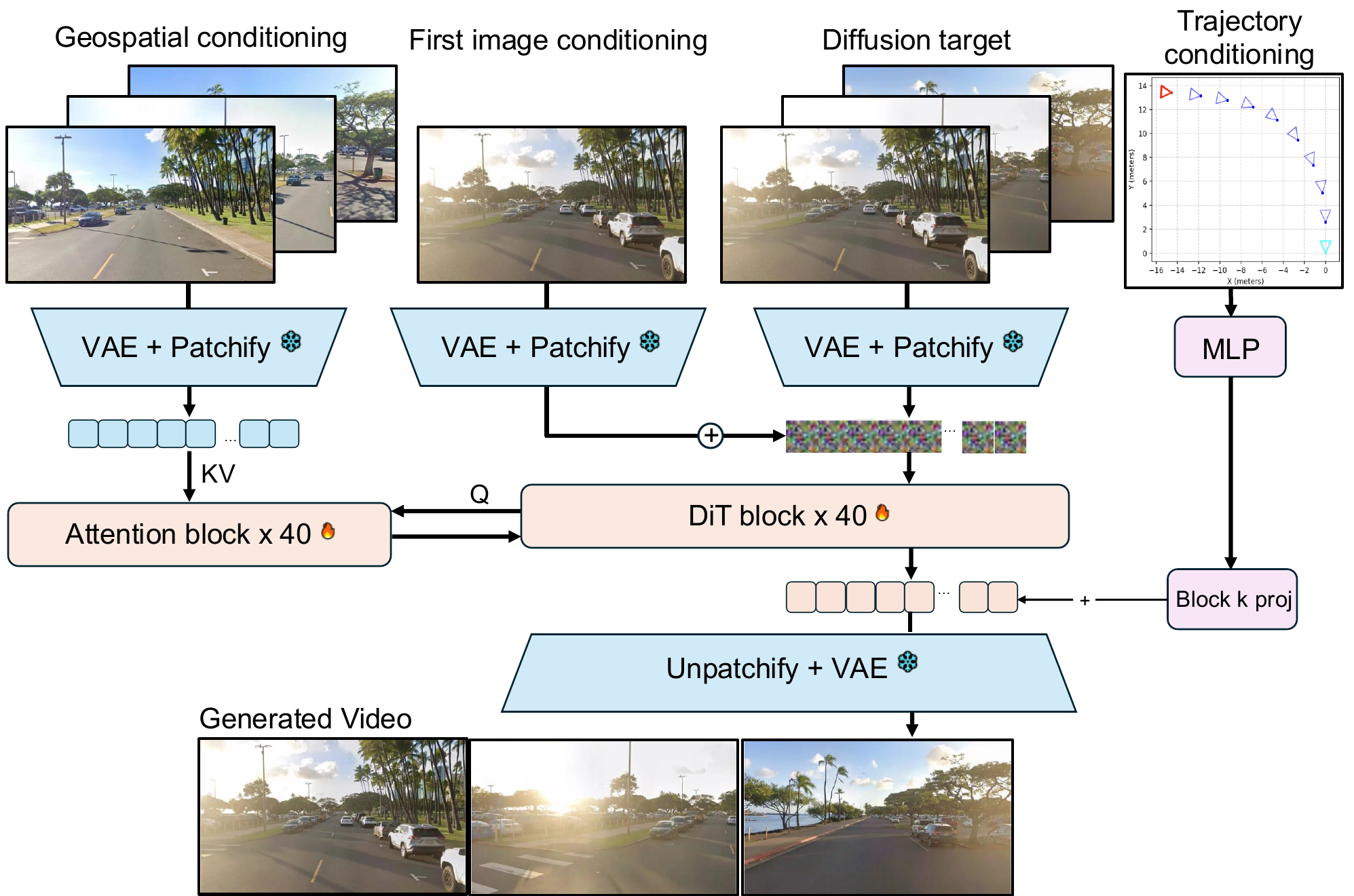}
 \caption{\textbf{Architecture.} The generator takes three conditions: the first image, a trajectory, and geo-registered videos (denoted geospatial conditioning) along the trajectory.}
\label{fig:arch}
\vspace{-2em}
\end{wrapfigure}

We finetune from a state-of-the-art image-to-video (I2V) generative model, Wan 2.1 (14B) \citep{wan2025}. It consists of a spatio-temporal VAE and a diffusion transformer (DiT). We refer readers to the original paper for details. 

Our goal is to condition the model on a first image that initializes the scene, a defined trajectory, and geo-registered data in the form of video frames along the defined trajectory. We visualize our architecture in \cref{fig:arch}.

\noindent\textbf{First image conditioning.} We follow the same conditioning in the I2V base model. The first image of the target video is independently processed by the VAE, then padded and concatenated channel-wise to the the target latents. This image initializes the scene: the generated video is expected to follow its lighting conditions and animate the dynamic objects in the frame.

\noindent\textbf{Trajectory conditioning.}
We specify a trajectory as a list of 4x4 extrinsic matrices. During training, we convert them to relative poses on-the-fly, where the first frame of the trajectory is at the origin. The poses are originally in ECEF coordinates, so the relative poses are in metric scale (meters) and consistent across cities.

We then flatten the matrices, downsample the temporal dimension by 4x with a Conv1D layer to match the temporal downsampling of the VAE, process them with a two-layer MLP, and then use a zero-initialized projection layer to match the dimensions of the Wan model, one for each attention block. The output of the $k$-th projection layer is added to the output of the $k$-th DiT block.
This allows pose information to be weighted depending on which block is more important for handling video movement, without disrupting the video prior.

However, one limitation of Street View data is that the majority of trajectories move straight along driving paths. To improve generalization, we augment rotations by cropping the panoramas at random yaws. Specifically, we randomly select a yaw between 0 and 360 degrees as the starting viewing angle, and add a rotation uniformly sampled between 0 and 2 degrees between each frame. 
We show that our model generalizes to out-of-distribution rotations in \cref{fig:ood_test}.

\noindent\textbf{Geospatial conditioning.}
We sample from the set of paired panoramas during training. One serves as the target video to generate, the other as context for grounding. We crop both to a fixed $65^{\circ}$ field of view (FOV), with the yaw following the rotation augmentation determined by the camera pose. 

Because these captures are separate traversals, they exhibit spatial and temporal discrepancies. Spatial shifts occur due to variations in camera centers (e.g., lane changes), while temporal misalignments result from varying vehicle speeds.
To ensure our model understands the relation between the two sets of captures and the underlying 3D scene, we vary the length of the condition panoramas during training. This forces the model to become robust to these discrepancies, rather than relying on a one-to-one mapping between frames. We show in \cref{fig:comparisons} that the generated sequences can contain accurate renderings of buildings that appear much later in the condition frames.

We pass the conditions to the Wan model via cross-attention.
We duplicate the original self-attention blocks from the pretrained base model, but train them separately (denoted Attention block in \cref{fig:arch}). During training, we first pass the condition video through the VAE, then use the latents as the keys and values for cross-attention. The target noisy latents serve as the query. This strategy allows each frame in the target sequence to attend to the entire context of the condition. 

Since cross-attention sequences can be varying length, we randomly set the number of conditioning frames to be between 61 and 81, such that the number is mismatched to that of target frames, to force the model to extract global context rather than only pixel-aligned correspondences.

\noindent\textbf{Classifier-free guidance} \citep{ho2022classifier}\textbf{.} We set the unconditional probability to 10\% for both the poses and panorama conditions, but sampled independently, so both can still be valid conditions in the absence of the other. Importantly, we find that without geospatial conditions, the model falls back to being a trajectory-conditioned I2V model, showing that it retains strong generative priors without only relying on external context. The left of \cref{fig:ood_test} shows such an example. The geospatial condition is completely mismatched to the trajectory due to traffic and therefore effectively ignored, yet the model follows the other conditions and generates a plausible video. 

\vspace{-1em}
\subsection{Inference via User Input and RAG} 
\label{sec:inference}

\begin{figure}[t!]
  \centering
  \includegraphics[width=\textwidth]{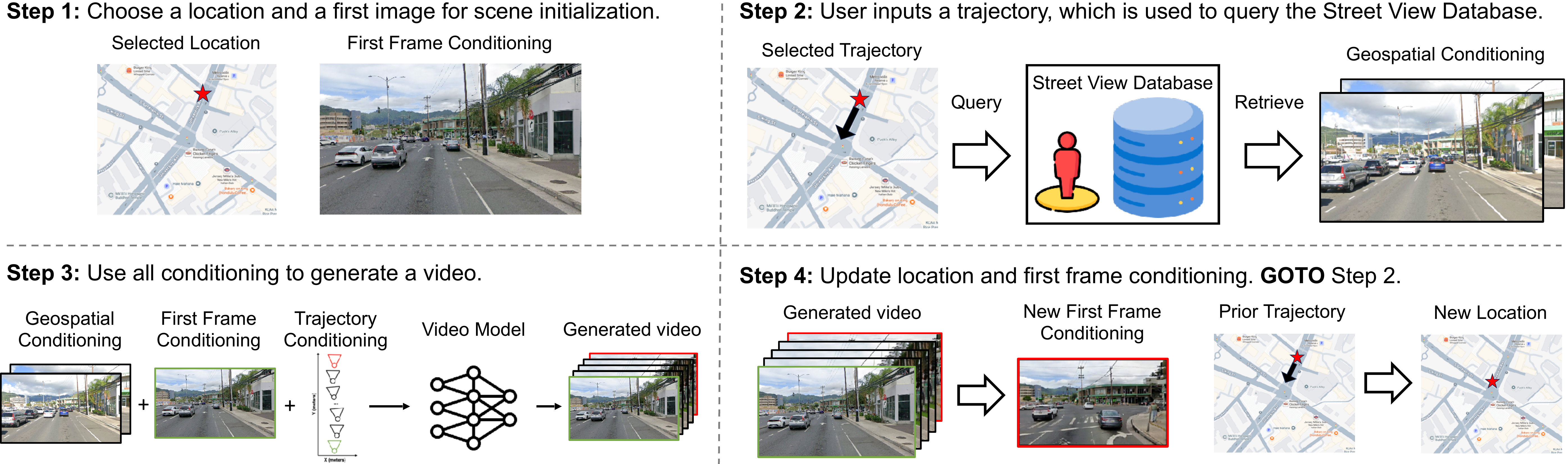}
  \caption{ \textbf{RAG pipeline at inference-time.}
  The user first selects a location and image that they want to step into. Then with a user-specified trajectory we use the Street View Database to retrieve our geospatial conditioning. All conditions are passed to the video model which generates the output the user sees. We then automatically update the first frame and location and repeat the process.
  }
  \label{fig:inference_rag}
\end{figure}

As shown in~\cref{fig:inference_rag}, we describe the full pipeline of generating a consistent, minutes-long video given a user-defined trajectory and first image condition. 

 \begin{figure}[t!]
  \centering
  \includegraphics[width=\textwidth]{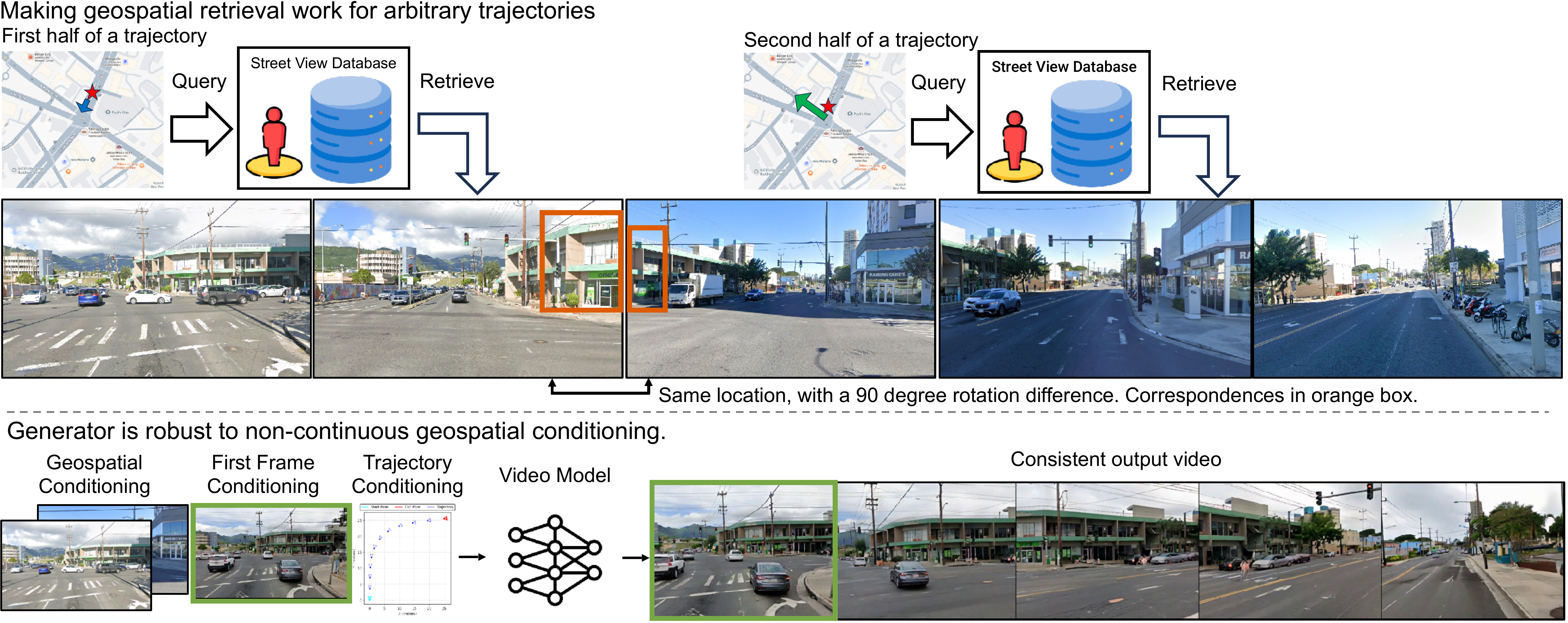}
  \caption{ \textbf{Making geospatial retrieval work for arbitrary trajectories.}
  Navigating arbitrary trajectories may require stitching together distinct videos from the database. In this example, since the initial retrieved path continues straight, CityRAG retrieves a second, perpendicular path from the same intersection to construct a new trajectory that resembles turning right at the intersection. Despite the discontinuity in the geospatial condition frames, the generator produces a consistent video, indicating its robustness and its understanding of the static and transient elements in a scene.
  }
  \label{fig:detailed_retrieve}
  \vspace{-1em}
\end{figure}

First, we randomly pick an image from the dataset, or a casual capture from the internet, or even a AI-modified image with snowy conditions (Honolulu scene in \cref{fig:teaser}). We identify the location on the map and ask the user for a trajectory (\cref{fig:inference_rag}, Steps 1 and 2). 
Next, we retrieve geo-registered Street View data along the defined path, and use them as conditioning to generate a video (Steps 2 and 3). Then, we can repeat this step autoregressively until we reach a desired location (Step 4). 

The defined path may not exist as a single geo-registered video, so we stitch frames from multiple videos to navigate arbitrary trajectories. As shown in \cref{fig:detailed_retrieve}, the initial retrieved path continues straight, so CityRAG retrieves a distinct video from cross traffic. By stitching the two videos, we create a proxy trajectory that turns right at the intersection (at a 90 degree angle). Though the model was always trained on continuous geospatial videos, the generated videos remain consistent with discontinuities in the conditions during testing. This indicates our model understands static and transient elements in a scene and is robust to appearance changes and pixel-mismatches in the conditions.

Our generated video of San Juan in \cref{fig:teaser} is conditioned on multiple (4) stitched geospatial videos, yet remains consistent across a thousand frames. We show more examples on our website. 


\section{Experiments}
\label{sec:exp}

\noindent\textbf{Baselines.}
To the best of our understanding, there are no open-source video generation models that perform our task, which requires navigation control (trajectory conditioning) and adherence to both an initial condition (first image) for scene appearance and an external condition (geospatial video) for spatial grounding.
We identify three closely related lines of work and run baselines from each:

\noindent1) I2V + pose control. We use Gen3C \citep{ren2025gen3c}, a state-of-the-art video model with camera control. It shows driving simulations as one of its applications.  

\noindent2) V2V + pose control. We use another variant of Gen3C and TrajectoryCrafter \citep{Yu_2025_ICCV}. Both methods take a dynamic input video and re-render it given a different trajectory. For our setup, we provide the conditioning frames and re-render with the target camera trajectory. 

\noindent3) V2V + style transfer. We use AnyV2V \citep{ku2024anyv2v}, a method that transforms a video to the style of an image. We provide the geospatial video as input, and the first image as the style reference. 

\noindent\textbf{Train-test split.} 
From the 10 cities in our dataset, we use the first 8 for both training and testing. For testing, we hold out entire neighborhoods to ensure samples come from streets unseen during training. 
We reserve the remaining two cities, London and São Paulo, only for testing. In short, none of the quantitative and qualitative results shown in the paper are of streets seen during training. We did not observe any differences in generated visual quality for held out sets of training cities and for completely withheld cities. See videos in the supplement for a visual comparison and \Cref{sec:quantitative} for metrics.

For testing, we filter for trajectories that feature at least a 45° rotation to avoid models relying solely on context from the first image. Then, from each city, we randomly select 10 trajectories. 
For comparing with baselines, we do not perform autoregressive generation or provide a user-defined trajectory, but use preprocessed pairs of trajectories, as described in \Cref{sec:data}.

\subsection{Qualitative Comparisons}

\begin{figure}[t!]
  \centering
  \includegraphics[width=\linewidth]{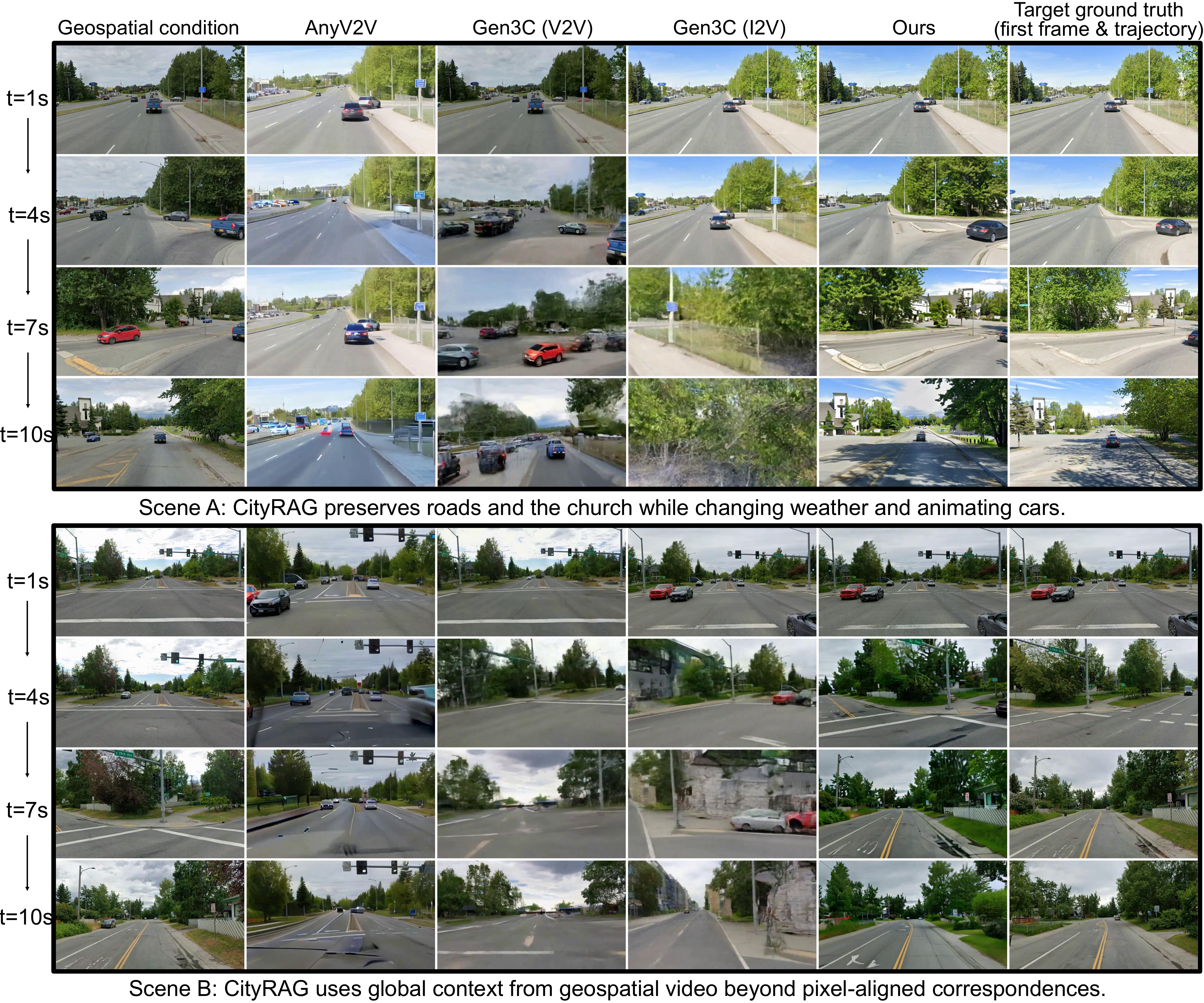}
  \caption{\textbf{Qualitative comparisons.} We show two challenging test samples. Input conditions include the video for geospatial conditioning (leftmost column) and the first image and the trajectory of the ground truth video (rightmost column). \textit{Scene A:} CityRAG follows the weather and animates the black car in the first image, and faithfully renders the church and roads. \textit{Scene B:} CityRAG renders the building and fence (t=7s) that appear later in the geospatial conditioning (t=10s, lags due to traffic), showing its ability to extract global context rather than only pixel-aligned details. 
  }
  \label{fig:comparisons}
  \vspace{-1em}
\end{figure}

In \Cref{fig:comparisons}, we highlight and analyze two challenging test samples, and show dozens of video results in the supplement. The video for geospatial conditioning (leftmost column), trajectory defined by the target video (rightmost column), and the first image of the target video are provided as conditions. 
CityRAG successfully handles these scenarios. In Scene A, 
the generated video follows both the weather conditions and the cars of the first image. As the video progresses, the black car in front continues to move realistically, and reappears even when it goes out of sight during the turn.

\begin{figure}[t]
  \centering
  \includegraphics[width=\textwidth]{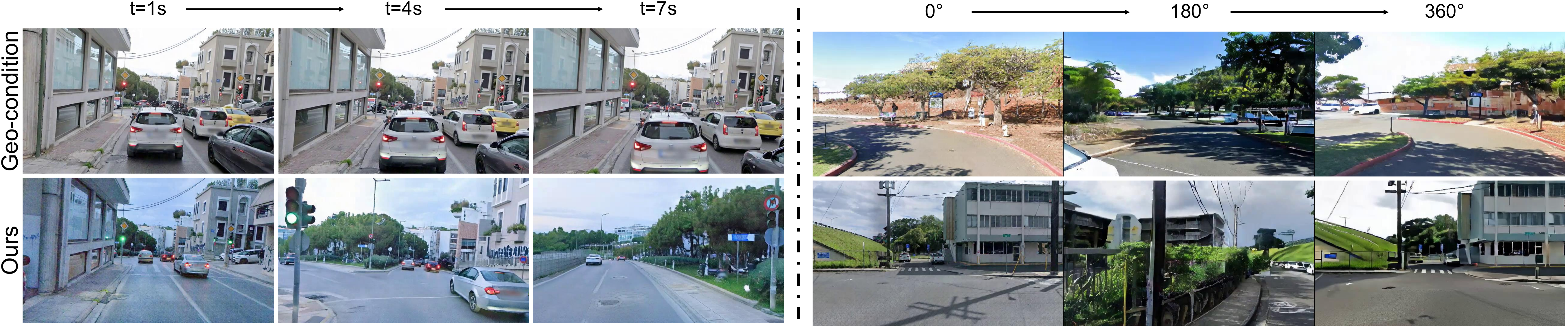}
  \caption{ \textbf{Flexibility of trajectory conditioning.} Our trajectory conditioning does not have to be precisely aligned with the geospatial conditioning.
  \textit{Left:} Even though there is a mismatch between the geospatial condition (car stuck in traffic) and trajectory (left turn), our model generates a plausible sequence following the trajectory. \textit{Right:} Our model can rotate 360° in a single sequence. The low visual quality is an artifact of the temporal VAE.
  }
  \label{fig:ood_test}
  \vspace{-1em}
\end{figure}

In Scene B, we show that CityRAG follows pose precisely even when there is a mismatch between it and the geospatial condition. Specifically, the geospatial condition stops at the intersection to yield to oncoming cars (see t=4s). However, the generated video follows the pose and accurately renders the structure at t=7s that only appears in the geo conditioning at t=10s. This shows our model can extract and render the structure of the scene, rather than relying on a pixel-aligned transfer as in V2V models. We also show a similar example in \Cref{fig:ood_test}. The capture representing the geospatial condition gets stuck in traffic, yet the model produces a plausible generation. 


All baselines fail. 
AnyV2V copies over the first image but fails to reconcile the differences between the source video (geospatial condition) and the first image. In both scenes, the virtual camera never moves. 
Gen3C (I2V) shows stability and relatively high visual quality in the first few seconds (t=1s to 4s) when the car is only moving forward, but its generation breaks down when the car turns. 
Gen3C (V2V) struggles even more with complex poses. Through testing, we find that it 
can only re-render videos with very limited camera movement (e.g., a small wobble). 

\noindent\textbf{Flexibility of trajectory conditioning.} 
We further demonstrate CityRAG's robustness and flexibility.
In \Cref{fig:ood_test}, the geospatial video is stuck in traffic. However, CityRAG's generation follows the defined trajectory to move forward and take a left turn, showing its strong generative priors even without external context. 
Our model 
can also follow extreme rotations, such as 360° within a single sequence, which is double the maximum rotation present in the training set.


\subsection{Quantitative Comparisons}
\label{sec:quantitative}

\noindent\textbf{Reconstruction and fidelity.}
A major objective of our task is fidelity to the ground truth scene. Thus, we first evaluate with metrics including PSNR, LPIPS \citep{zhang2018perceptual}, and SSIM \citep{wang2004ssim} (\Cref{tab:nvs}). Since we are focused on static structures, we also evaluate on a static-variant of these metrics (denoted -S). Specifically, we use Mask2Former \citep{cheng2021mask2former} to segment dynamic classes (i.e., vehicles and people), and mask their pixels during calculation. We also include FID \citep{heusel2017gansfid} that assesses the quality of generated images by comparing their feature distributions to those of real images. Lower indicates generations are more similar to real images.

Compared to dedicated view synthesis or reconstruction techniques, all methods we test obtain relatively low scores, including ours. This is because generative models are inherently stochastic and do not aim for the exact pixel-level reconstruction or overfitting that traditional NVS methods prioritize. 
Minor shifts in camera movement or hallucination of plausible geometry can lead to high pixel-wise error. 
Despite this, CityRAG outperforms all baselines and maintains the best fidelity to the ground truth scenes, which matches our qualitative observations. Furthermore, our method significantly leads in metrics that measure perceptual similarity, such as LPIPS and FID.
 
We also note that we did not observe any meaningful performance gap between the held out scenes of the eight trained cities and the two untrained cities, suggesting that our method is generalizable to a variety of diverse scenes and conditions. Specifically, the PSNR, SSIM, LPIPS, and FID scores of the generations of the untrained cities were 15.11, 0.461, 0.517, 16.90, respectively, comparable to that of the full test set.

\clearpage\noindent\textbf{Trajectory adherence.}
We evaluate how closely generated videos follow the reference camera trajectory. We recover camera poses from both the generated and ground truth videos using MegaSaM \citep{li2025megasam}, and align each generated trajectory to its reference with a single $\mathrm{Sim}(3)$ transformation, accounting for rotation, translation, and scale. We report absolute trajectory error (ATE) as the root mean squared error (RMSE) of camera positions, and rotation RMSE as the error in camera orientations at corresponding frames. We also report relative pose error (RPE) in translation and rotation over one-second intervals, measuring discrepancies in local camera motion. Translation errors are expressed as percentages of the reference trajectory's total path length, and rotation errors are measured in degrees. We average the results over all 100 test scenes; lower is better for all four measures. As shown in \Cref{tab:grounding}, CityRAG achieves the lowest errors throughout, with an ATE of 2.49\% and rotation RMSE of $6.32^{\circ}$, compared to 7.78\% and $57.88^{\circ}$ for Gen3C V2V. These results agree with the qualitative observations that CityRAG follows trajectories with substantial camera rotation.

\noindent\textbf{Static-layout consistency.}
We measure the similarity of static scene content after accounting for viewpoint differences using reprojected DINOv2 \citep{oquab2023dinov2} features. We segment generated and ground truth images with OneFormer Swin-L \citep{jain2023oneformer}, retaining patches containing only road, sidewalk, building, wall, fence, pole, traffic light/sign, or terrain classes, and excluding vegetation, sky, people, and vehicles. Using MegaSaM depth and camera poses, we align each generated sequence to the ground truth with a single $\mathrm{Sim}(3)$ transformation, lift generated-image DINOv2 patch centers into 3D, and project them into the corresponding ground truth camera. We retain correspondences only when both patches are static, rejecting out-of-bounds projections, semantic boundaries, and depth discontinuities. We compute the cosine similarity between matched patch features and report the mean static DINO score across all 100 test scenes; higher is better. CityRAG achieves 0.611, compared to 0.370 for the strongest baseline, Gen3C V2V, indicating greater consistency with the static layout of the reference scene.

\begin{table}[t!]
  \caption{\textbf{Reconstruction and fidelity.} We calculate view synthesis metrics to measure the fidelity of generated videos to real-world scenes, and FID to measure visual quality.
  }
  \label{tab:nvs}
  \centering
  \scalebox{0.8}{
      \begin{tabular}{lccccccc}
        \toprule
        Method & PSNR ↑ & SSIM ↑ & LPIPS ↓ & PSNR-S ↑ & SSIM-S ↑ & LPIPS-S ↓ & FID ↓ \\
        \midrule
        TrajCrafter  & 11.90 & 0.403 & 0.705 & 11.92 & 0.536 & 0.548 & 55.45\\
        AnyV2V       & 11.82 & 0.385 & 0.698 & 11.83 & 0.521 & 0.551 & 47.56 \\
        Gen3C V2V    & 12.34 & 0.432 & 0.677 & 12.36 & 0.538 & 0.558 & 57.13\\
        Gen3C I2V    & 13.28 & 0.453 & 0.654 & 12.86 & 0.545 & 0.543 & 61.07\\
        Ours         &  \textbf{15.03} & \textbf{0.466} & \textbf{0.504} & \textbf{15.86} & \textbf{0.560} & \textbf{0.432} & \textbf{16.55}\\
      \bottomrule
\end{tabular}
  }
\end{table}

\begin{table}[t!]
  \caption{\textbf{Trajectory adherence and static-layout consistency.} Results are averaged over all 100 test scenes. Translation errors are normalized by reference path length; rotation errors are in degrees. RPE uses one-second intervals. Static DINO measures cosine similarity of reprojected DINOv2 features in static regions.}
    \label{tab:grounding}
      \centering
        \small
          \setlength{\tabcolsep}{4pt}
            \begin{tabular}{lccccc}
                \toprule
                    & \multicolumn{4}{c}{Trajectory adherence} & Static layout \\
                        \cmidrule(lr){2-5}\cmidrule(lr){6-6}
                            Method & \shortstack{ATE RMSE $\downarrow$\\(\% path)} & \shortstack{Rot. RMSE $\downarrow$\\($^{\circ}$)} & \shortstack{RPE trans. $\downarrow$\\(\% path)} & \shortstack{RPE rot. $\downarrow$\\($^{\circ}$)} & Static DINO $\uparrow$ \\
                                \midrule
                                    TrajCrafter & 13.69 & 102.60 & 13.14 & 10.21 & 0.339 \\
                                        AnyV2V & 22.44 & 128.64 & 17.29 & 11.22 & 0.240 \\
                                            Gen3C V2V & 7.78 & 57.88 & 7.92 & 4.55 & 0.370 \\
                                                Gen3C I2V & 9.45 & 82.89 & 10.92 & 6.55 & 0.299 \\
                                                    CityRAG & \textbf{2.49} & \textbf{6.32} & \textbf{1.95} & \textbf{3.84} & \textbf{0.611} \\
                                                        \bottomrule
                                                          \end{tabular}
                                                          \end{table}
                                                          \clearpage\subsection{User Study}
\begin{wrapfigure}{r}{0.5\textwidth} 
\vspace{-1em}
\includegraphics[width=0.5\textwidth]{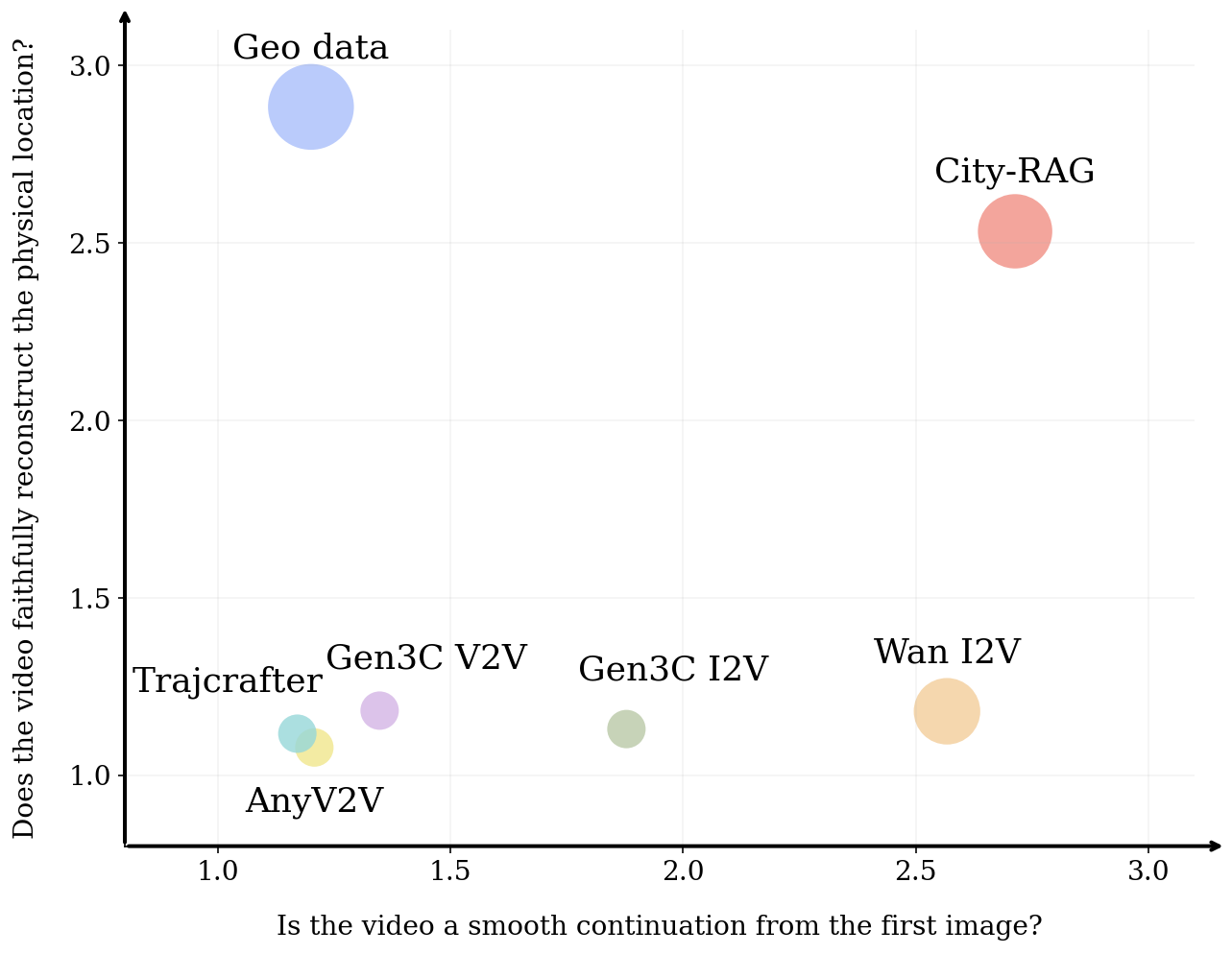}
 \caption{\textbf{User study results.} Users rate each video on a scale of 1 (lowest) to 3 (highest). 
 Only CityRAG generates videos that are both smooth continuations from first images and faithful renders of the real physical location.}
\label{fig:userstudy}
\vspace{-3.9em}
\end{wrapfigure}
We conduct a user study consisting of three questions, asking users to evaluate 1) visual quality, 2) whether videos are smooth continuations of the first image, and 3) fidelity to the physical location (using the last frame of the geospatial conditioning video as the reference destination). Users rate each sample 1 (lowest), 2, or 3 (highest). 
We provide details in the supplement. 

The responses are plotted in \Cref{fig:userstudy}. The $x$- and $y$-axes record the average scores for the second and third questions, respectively, and the radius indicates visual quality; larger is higher. In addition to our baselines, we include the Wan I2V base model and the retrieved geo-registered data as a reference for the specific axis each specializes in. Again, we observe that our task is a novel yet practical problem setting that existing methods cannot perform.

\subsection{Discussion}
We acknowledge that the baselines are not trained for our task, and therefore expectedly fail. The closest approach, conceptually, would be methods like AnyV2V that perform style transfer on a video; i.e., transform an input video (geospatial condition) into a desired appearance (first image). But this transfer is extremely non-trivial. The model would need to understand that the cars, pedestrians, and weather are all part of the ``style.'' It would then need to have the flexibility to animate the cars and pedestrians realistically.
And fundamentally, the trajectory of the generated video could only follow that of the input video, rather than taking user-defined trajectories. 

Thus, we highlight the robustness and flexibility of our approach: CityRAG creates a simulated environment of a scene by understanding the global structure even when geospatial conditions are imperfect, realistically preserves and animates the scene initialization, and allows users to freely navigate arbitrary trajectories.


\section{Conclusion}
\label{sec:discussion}

To the best of our knowledge, CityRAG is the first video generative model that emphasizes adherence to our real world, and therefore it can help unlock a variety of applications that rely on specific environment layouts. Furthermore, we introduce a fully data-driven strategy that teaches the model to be aware of semantics, generalizable to diverse conditions, and to disentangle static and dynamic attributes via temporally unaligned training data.

In the appendix, we provide additional details including (A) Limitations and future work, (B) Latency and inference costs, (C) Training optimization, (D) Architecture ablations, (E) User study details, (F) Extended related works, (G) Ethics and privacy.
We also encourage readers to view video results hosted on our \href{https://cityrag.github.io/}{\textcolor{blue}{\underline{website}}}.




\section*{Acknowledgements}
Gene Chou was supported by an NSF graduate fellowship (2139899).
We thank Gordon Wetzstein, Aleksander Holynski, Jon Barron, Dor Verbin, Pratul Srinivasan, Rundi Wu, Ruiqi Gao, and Haofei Xu for discussions and support.

\bibliographystyle{plainnat}
\bibliography{references}

\appendix

\section{Limitations and Future Work}
There are a few limitations of CityRAG. First, we perform autoregression by only providing the generated last frame as the first frame of the subsequent sample. Existing methods for autoregression could be incorporated to improve long-term consistency.
However, we do want to highlight the stability that geospatial grounding brings. Using the last frame as the first image of the next generation leads to significant drift and degradation after just 1 or 2 iterations in typical I2V models, yet our generated scenes remain stable (especially the static structures) even after dozens of iterations. We believe that this technique would complement existing autoregression methods.

Second, we provide no heuristics to the model regarding static vs. transient objects — the disentanglement is completely data-driven. Future work could include fine-grained control and annotations over individual elements in the scene to improve controllability and customization.

Third, though we mention potential downstream applications such as virtual tourism and simulations, our method is not real-time. A truly interactive video world model requires multiple dimensions of progress, including faster inference, long-horizon temporal consistency, realism, controllability, and more. CityRAG specifically focuses on spatial grounding to real physical locations for realism and long-horizon temporal consistency, and is orthogonal to other dimensions. We describe our inference costs and latency in the next section.

Fourth, because we do not have captions for our data, we freeze the original text cross-attention blocks and use a fixed prompt: ``A photorealistic, cinematic video of a city street. The camera performs a smooth, steady tracking shot moving along the asphalt road, maintaining a consistent level angle that offers an immersive street-level perspective.'' After finetuning, we observe the model no longer responds to new text captions; we leave captioning Street View and text conditioning for future work.

Finally, there exist data biases. For instance, the data does not include snowy, rainy, or nighttime conditions due to hardware and sensor limitations. Augmenting the data, and perhaps introducing modalities like text, is another important future work. 

\section{Latency and Inference Costs}

During training and inference, the target video consists of 73 frames at 480p resolution ($832 \times 480$). The VAE downsamples the temporal dimension by $4\times$ and spatial dimensions by $8\times$. With a DiT patch size of 2, the resulting latent size is $18 \times 30 \times 52$ ($T \times H \times W$).

The inference costs can be divided into two parts: retrieving the geospatial conditions from the Street View database, and the video generation. We first built the entire map of a city using Scipy’s cKDTree, such that given any image with GPS information, the lookup time for the nearest neighbor is O(log N). This is a one time cost and less than a minute in wall time. All sequential frames following the desired relative poses can be retrieved in O(1) constant time. 

The video generation cost is comparable to the Wan 2.1 base model. We used multiple inference tricks such as TeaCache \citep{liu2024timestep} and DeepSpeed-Ulysses \citep{jacobs2023deepspeed}. Inference is 40 steps using the UniPC \citep{zhao2023unipc} solver.

On a node of 8 A100 GPUs, the inference wall time measured 90 seconds for 73 frames, with an amortized cost of roughly 0.8 FPS.
On a node of 8 B200s, the time was reduced to 30.5 seconds, or 2.4 FPS. 

As mentioned in the previous section, faster inference was not the focus of this project, but we believe ongoing works in distillation, few-step generation, caching...etc will continue to improve latency. For instance, a model distilled to 4 steps would roughly reduce the inference time 10 fold.

\section{Training Optimization} 
We adopt the v-prediction \citep{salimans2022progressive} objective with a shifted noise schedule toward higher timesteps (a factor of 3.0, following SD3 \citep{esser2024scaling} and Flux \citep{labs2025flux1kontextflowmatching}). We use the Muon \citep{liu2025muonscalablellmtraining} optimizer with a fixed learning rate of 1e-5 with warmup. We train our model on 32 A100 GPUs for a week, for 20k iterations. 
Empirically, the AdamW \citep{kingma2015adam} optimizer required significant noise schedule shift toward higher timesteps ($t>900$), which led to a degradation in output visual quality. 
We also implement FSDP and gradient checkpointing to reduce vram usage.

\section{Architecture Ablations}
Due to computation constraints, we did not run all design ablations on the entire training set. Here, we provide quantitative results when training and testing only on one city (Honolulu), with the testing set streets of a few unseen neighborhoods, same as the setup described in the main paper.

\begin{table}[t!]
  \centering
  \caption{Ablations for trajectory conditioning (left) and geospatial conditioning (right).}
  \label{tab:ablation}

  \begin{minipage}[t]{0.49\linewidth}
    \centering
    \resizebox{\linewidth}{!}{
      \begin{tabular}{lcccc}
        \toprule
        Method & PSNR ↑ & SSIM ↑ & LPIPS ↓ & FID ↓ \\
        \midrule
        Plucker Rays & 13.39 & 0.432 & 0.503 & 15.51 \\
        Concat MLP Output & 13.31 & 0.422 & 0.516 & 14.82 \\
        Residual Add (Ours) & \textbf{16.40} & \textbf{0.486} & \textbf{0.485} & \textbf{14.50} \\
        \bottomrule
      \end{tabular}
    }
  \end{minipage}
  \hfill
  \begin{minipage}[t]{0.49\linewidth}
    \centering
    \resizebox{\linewidth}{!}{
      \begin{tabular}{lcccc}
        \toprule
        Method & PSNR ↑ & SSIM ↑ & LPIPS ↓ & FID ↓ \\
        \midrule
        No Geospatial & 14.13 & 0.479 & 0.505 & \textbf{14.04} \\
        VGGT + CrossAttn & 14.11 & 0.479 & 0.502 & 14.07 \\
        RGB + ControlNet & 16.01 & 0.484 & 0.499 & 15.28 \\
        RGB + CrossAttn (Ours) & \textbf{16.40} & \textbf{0.486} & \textbf{0.485} & 14.50\\
        \bottomrule
      \end{tabular}
    }
  \end{minipage}
\end{table}
\noindent\textbf{Trajectory conditioning.}
We also experimented with other conditioning methods: 1) concatenating plucker rays to the noisy target latents, which is common in novel view synthesis methods; 2) concatenating the MLP output (same process as residual addition described in the paper) to the noisy target latents by repeating the values across the spatial dimensions. We empirically found that our method of residual addition worked best. Metrics on Honolulu are provided in \cref{tab:ablation}. The first image and geospatial conditions are the same as those described in the main paper.

\noindent\textbf{Geospatial conditioning.}
We explored alternative methods for formatting the conditioning video, such as directly using the raw panoramas.
While this alternative should provide a complete 360-degree view of the scene and thus more context for conditioning, we found that the model struggled to follow the context, likely because this data was rare or absent in the Wan base training. Even with extensive training, the model showed no signs of rendering buildings and roads outside the first image's FOV. 

Before using RGB images as conditions, we also tried VGGT \citep{wang2025vggt} features, which has, in many applications, demonstrated 3D awareness and semantic understanding of static and dynamic elements. However, our model showed no signs of integrating this information, despite various adapter layers, prolonged training, or even during small-scale, overfitting experiments. This observation is supported by concurrent work such as Gen3R \citep{huang2026gen3r3dscenegeneration}, which only successfully injected VGGT features into Wan after training an adapter to recast the features into the distribution of the VAE latents via KL and reconstruction losses.

In short, we found that the Wan base model more effectively integrates information that originate from its own temporal VAE. Shifts in the latent distribution hinder the learning process. For instance, in an ablation where condition frames were processed individually through the VAE (treating them as independent image latents rather than a video latent) the model required three times as many iterations to converge.

Finally, we conducted experiments with other conditioning mechanisms such as ControlNet \citep{zhang2023adding}, and found that our approach yielded the best adherence to the conditions. Metrics on Honolulu are provided in \cref{tab:ablation}. The first image and trajectory conditions are the same as those described in the main paper. Using VGGT + CrossAttn effectively kept the model a trajectory-conditioned I2V model, with no additional context learned by the base model. Note that with ControlNet, we set the first 10 layers as the encoder from which we cloned weights, and needed to continue to finetune the remaining 30 (decoder) layers. The main potential benefit of this architecture was the skip connections, though our final proposed method of cross attention worked best.

\begin{figure}[t]
  \centering
  \includegraphics[width=0.32\textwidth]{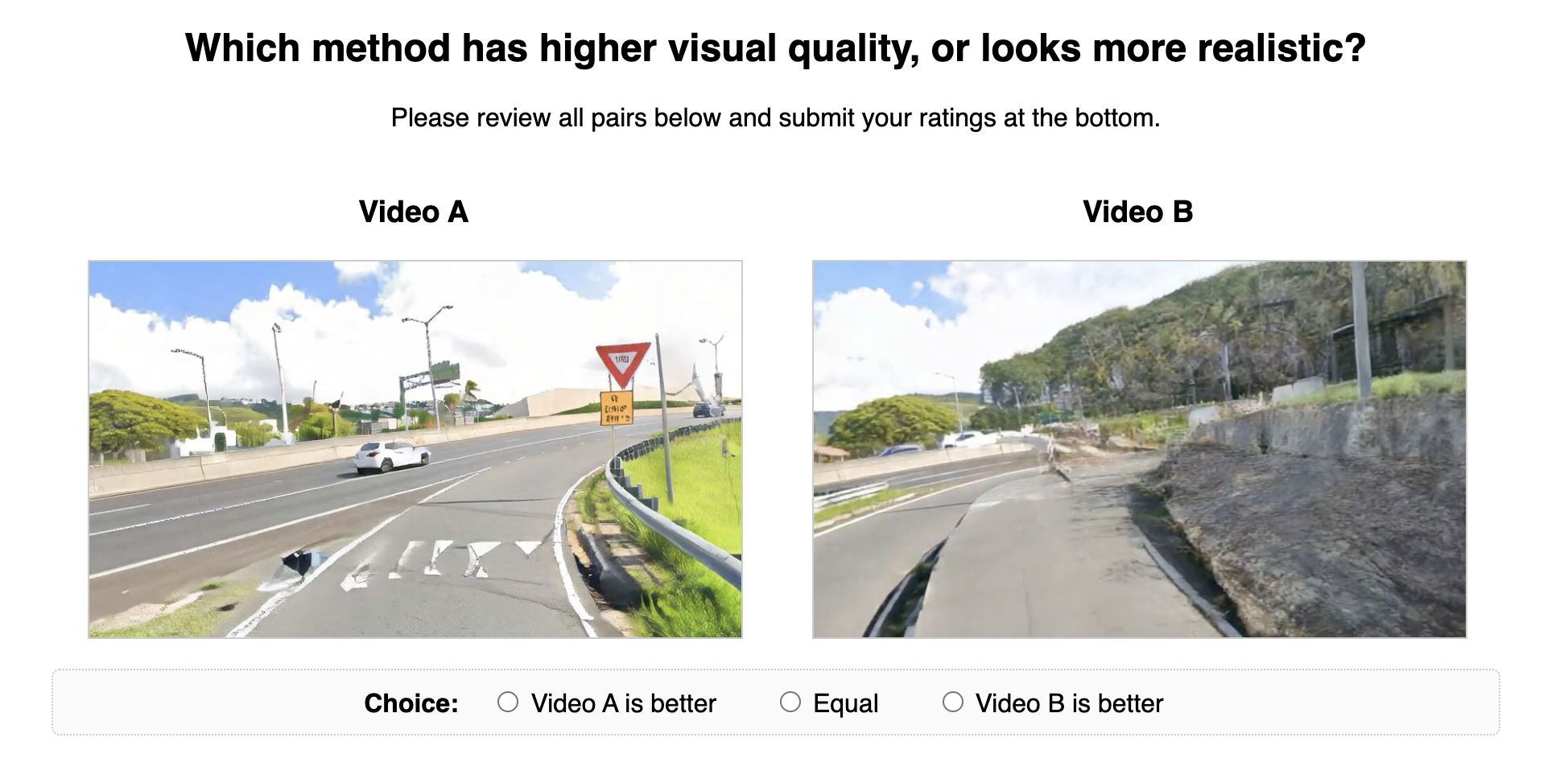}
  \hfill
  \includegraphics[width=0.32\textwidth]{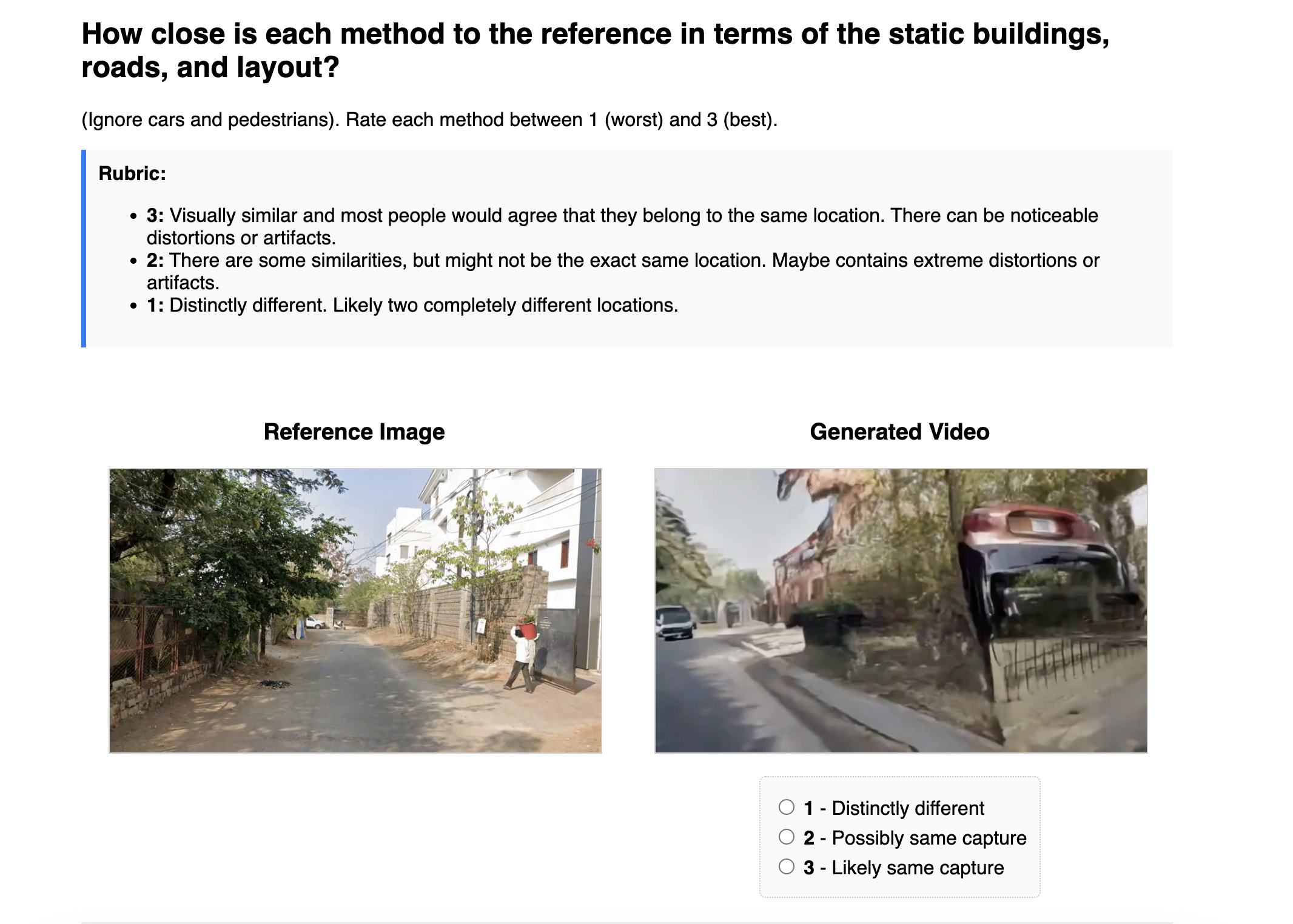}
  \hfill
  \includegraphics[width=0.32\textwidth]{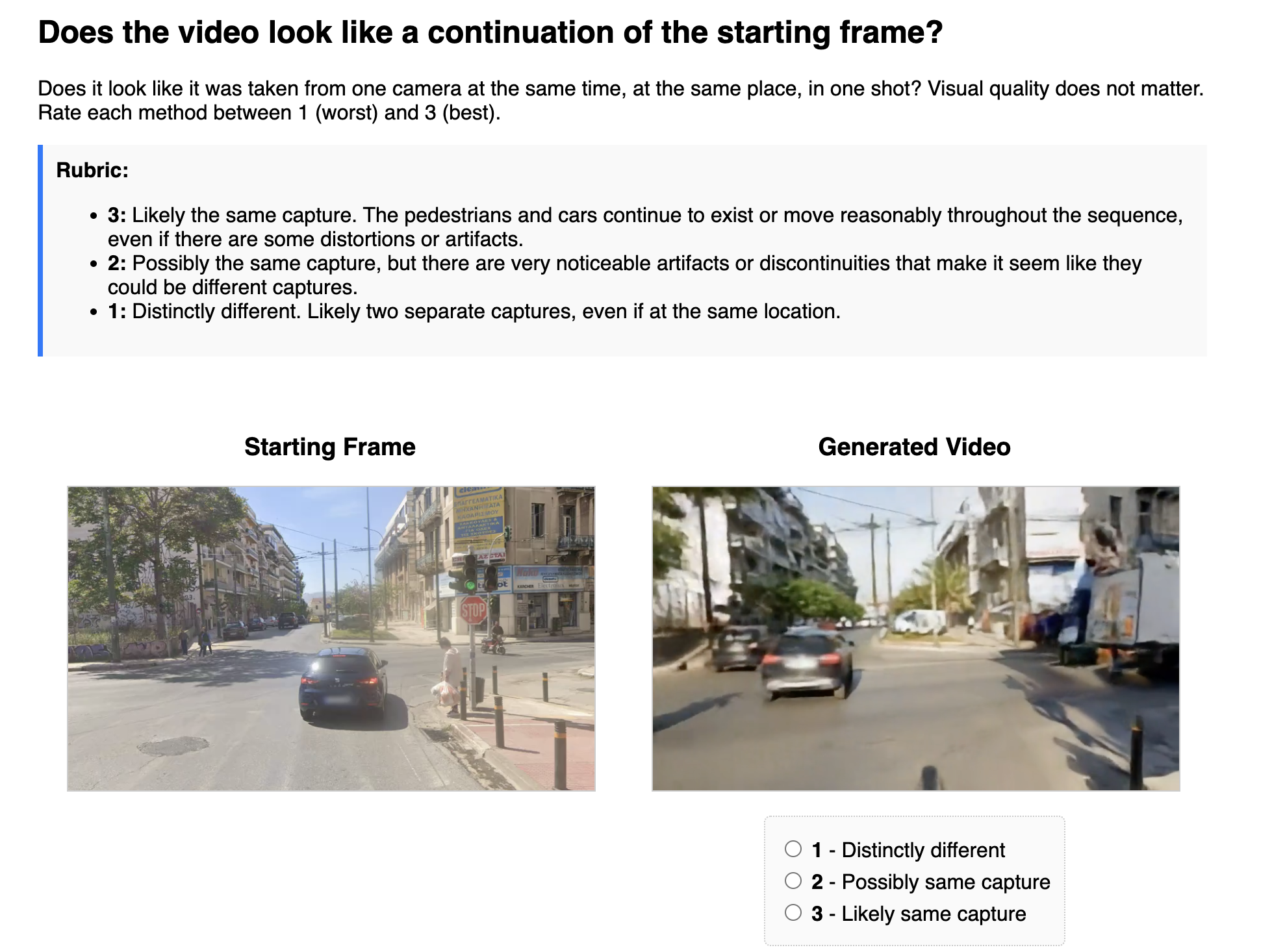}

  \caption{Three types of questions in our user study.}
  \label{fig:user_study_interface}
\end{figure}

\section{User Study Details}

As mentioned in the main text, we conduct a user study to evaluate the capabilities and limitations of each method. We set up three questions with 10 samples each, randomly sampled from our evaluation set. In total, we collected responses from 20 users. We provide screenshots of the interface in \cref{fig:user_study_interface} The questions are as follows:

\medskip\noindent\textbf{Q1: ``Which method has higher visual quality, or looks more realistic?''} We provide two videos, A and B, and three choices: ``Video A is better,'' ``Equal,'' ``Video B is better,'' and we conduct a head-to-head between our method and all baselines.

\medskip\noindent\textbf{Q2: ``Does the video look like a continuation of the starting frame? Does it look like it was taken from one camera at the same time, at one place, in one shot? Visual quality does not matter. Rate each method between 1 (worst) and 3 (best).''} We provide a starting frame, which is the first image condition explained in the main text. Then, we provide a generated video from a random method, and ask users to rate based on the following rubric.

\noindent3: Likely the same capture. The pedestrians and cars continue to exist or move reasonably throughout the sequence, even if there are some distortions or artifacts.

\noindent2: Possibly the same capture, but there are very noticeable artifacts or discontinuities that make it seem like they could be different captures.

\noindent1: Distinctly different. Likely two separate captures, even if at the same location.

\medskip\noindent\textbf{Q3: ``How close is each method to the reference in terms of the static buildings, roads, and layout? Ignore cars and pedestrians. Rate each method between 1 (worst) and 3 (best).''} We provide a reference image, which is the last image of the geospatial conditions and meant to signify the desired destination of the generated video. Then, we provide a generated video from a random method, and ask users to rate based on the following rubric.

\noindent3: Visually similar and most people would agree that they belong to the same location. There can be noticeable distortions or artifacts.

\noindent2: There are some similarities, but might not be the same location. Maybe contains distortions or artifacts. 

\noindent1: Distinctly different. Likely two completely different locations.

\section{Extended Related Works}
We further provide additional related work that is relevant but not central to the main paper.

\subsection{Driving simulations.} 
Although our work is not aimed to specifically address driving simulations, our training and evaluation domain is closely related. However, to the best of our knowledge, existing works have a different focus from ours. The simulations either look synthetic \citep{chen2026dwd, Chigot_2025, zhou2024simgen}, or cannot handle transfer of style, weather, and dynamic objects at once \citep{Yang_2025_ICCV, Ljungberphetal2025, deng2024streetscapes}.

\subsection{Large-Scale Novel View Synthesis and Reconstruction}
Novel view synthesis and reconstruction at city-scale are relevant to our task.
Early city-scale reconstruction were based on structure-from-motion (SfM) \citep{schoenberger2016sfm, Snavely2006PhotoTE}. For instance, Building Rome in a Day \citep{agarwal2009rome} handled 100K+ images via scalable SfM pipelines and cluster computing. These works established the foundation of large-scale reconstruction but focused on geometry rather than rendering.

Block-NeRF \citep{blocknerf} scaled NeRFs \citep{mildenhall2021nerf} to the city level by spatially decomposing scenes into many per-block NeRFs with appearance embeddings, pose refinement, and exposure alignment. Mega-NeRF \citep{turki2022meganerf} similarly partitions large outdoor regions into spatial cells and trains sub-modules with geometry-aware sampling to enable interactive flythroughs over areas orders of magnitude larger than single-scene NeRFs. Grendel-GS \citep{zhao2024scaling3dgaussiansplatting} distributes tens of millions of Gaussians \citep{kerbl20233d-3dgs} across GPUs to represent large scenes. 
However, these methods do not support dynamic motion and also require dense data with the same appearance because they use a deterministic rendering loss, which is difficult to obtain at scale.

\section{Ethics and Privacy}
As CityRAG aims to generate realistic videos of our world and is trained on a large corpora of Street View data, which in itself presents significant privacy and ethical challenges, it introduces unique challenges. 

\subsection{Privacy and Anonymization}
All of our data, prior to collection from the Street View database, were rigorously cleaned for identifiable information. All license plates and faces were blurred. Buildings and streets were blurred on request. No authors of this paper had access to the raw imagery.

Additionally, we heavily mitigated the appearance of people in the presentation of our results. We used tools such as Nano Banana to replace people in the condition images (both the first image and geospatial conditions) for synthetic ones, where applicable. We will also mask all people via a segmentation model when we show geospatial videos for the public release. 

We acknowledge these steps still cannot remove sensitive information 100\%, so we will closely monitor any request to remove videos and results after release. 

\subsection{Bias in Data Distribution}
Although we collected data from 10 cities, across 4 continents, the majority of the data is located in Western countries. This could introduce representation bias. Though CityRAG is a research paper without direct use in products or applications, in the future, any follow up work should attempt to mitigate this bias via more diverse data collection or algorithmic corrections. 


\end{document}